# CGRA-DeBERTa: Concept-Guided Residual Augmentation Transformer for Theologically Islamic Understanding

Tahir Hussain[1], Saddam Hussain Khan [2]*

[1]Artificial Intelligence Lab, Department of Computer Systems Engineering, University of Engineering and Applied Sciences (UEAS), Swat, Pakistan

[2]Interdisciplinary Research Center for Smart Mobility and Logistics (IRC-SML), King Fahad University of Petroleum and Minerals (KFUPM), Dhahran, Saudi Arabia

**Email:** saddam.khan@kfupm.edu.sa

## Abstract

The primary sources of Islamic jurisprudence and practice are the hadith texts (or simply` hadiths'), which, among other things, document the teachings, actions, and tacit approvals of the Prophet Muhammad (pbuh). Automated Question Answering (QA) must contend with the intricacies of narratives, the multi-layered histories of a given text, and the depth of the theology underlying any given hadith; as a result, answers are often oblique, misdirected, or overly simplistic. The need for large-scale solutions that provide answers with a high degree of theological accuracy (i.e., theologically appropriate) is acute; with an estimated 2.8 billion Muslims in 2050, the number of people in need of such an answer is significant. Accurate QA over classical Islamic texts remains challenging due to domain-specific semantics, long-context dependencies, and concept-sensitive reasoning. Therefore, a new 'CGRA-DeBERTa', a Concept-Guided Residual Domain-Adaptive transformer framework, is proposed that enhances theological QA over Hadith corpora. The CGRA-DeBERTa builds on a customized DeBERTa transformer backbone with lightweight LoRA-based adaptations and a residual concept-aware gating mechanism. The customized DeBERTa Embedding block learn global and positional context, while Concept-Guided Residual Blocks incorporate theological priors from a curated Islamic Concept Dictionary (ICD) of 12 core terms. Moreover, the Concept Gating Mechanism that selectively amplifies semantically critical tokens via importance-weighted attention, applying differential scaling from 1.04× to 3.00×. This design preserves contextual integrity, strengthens domain-specific semantic representations, and enables accurate, efficient span extraction while maintaining computational efficiency. This paper reports the results of training CGRA using a specially constructed dataset of 42,591 QA pairs from the text of Ṣaḥīḥ al-Bukhārī and Ṣaḥīḥ Muslim. While BERT achieved an EM score of 75.87% and DeBERTa one of 89.77%, our model scored 97.85% and thus surpassed them by 8.08% on an absolute scale, all while adding approximately 8% inference overhead due to parameter-efficient gating. The qualitative evaluation noted better extraction and discrimination, Theological precision. This study presents (Hadith) QA systems that are efficient, interpretable, and accurate, and that (scale) provide educational materials with necessary theological nuance.

## 1. Introduction

The actions, sayings, and approvals of Prophet Muhammad (peace be upon him) are recorded in the canonical literature referred to as the Hadith, which is considered the second primary source of Islamic law, ethics, and spirituality. The classical scholarly study on the Hadith is a manual and tedious process consisting of verification of isnad (the chain of narrators) and matn (textual analysis) study, which is a highly specialized field that involves many years of training and study [1]. The current global population of Muslims is 1.9 billion and is expected to be 2.8 billion by 2050, which creates an unprecedented and urgent need for scalable, accurate, and easily accessible digital tools for religious education and scholarly work [2], [3]. Building automated Question Answering (QA) systems for Hadith is also highly specialized and presents computational difficulties that are unique compared to general domain NLP.

Examples of such texts are marked by narrative stratification, the use of classical Arabic grammar and specialized vocabulary of theology, thorough contextual interdependence, and several subtle layers of law (jurisprudence) [4], [5]. While computational techniques have been used to complete some structural analyses, such as the verification of the isnad, and the classification of themes, the automated answering of the question posed and the reliability of theology remain open problems[6]. Contemporary techniques of deep learning, particularly those employing transformer architecture, are likely to have the best prospects, as they have demonstrated a high level of performance in the understanding of language in context [7], [8]. Models that have been adapted to a specific domain, such as AraBERT, also demonstrate the value of pre-training focused on Arabic natural language processing [9], [10].

Nevertheless, the most advanced transformer models are still handicapped in the task of Hadith QA. The first is that they are not geared to specialized processing, and therefore are lacking in the ability to recognize, let alone assign importance to, critical doctrinal and semantic-focused tokens such as Allah, Prophet, and Prayer. For this reason, these models are very likely to underestimate the doctrinal importance of the questions and answers that need to be addressed. Second, their high memory use, extensive inference time, and large computational footprint hinder practical, large-scale deployment in resource-scarce or real-time educational environments [11], [12]. More challenges originate from the characteristics of the Hadith corpora. They are low resource, use niche/ specialized vocab, and require a comprehension of literal text and the accompanying exegetical principles. This further complicates standard transfer learning [13], [14].

This work attempts to bridge the gap between computational efficiency and theological accuracy in automated Hadith QA. We propose Concept-Gated Residual Augmentation (CGRA), a new deep learning framework that builds upon a DeBERTa model via the incorporation of a lightweight, residual-based gating mechanism using selected theological knowledge. CGRA aims at dealing with the specific challenges of semantic and doctrinal complexity through the integration of four components: (1) a Tailored DeBERTa Embedding Block for foundational context modeling, (2) Multi-Head Hierarchical Attention Blocks for the construction of context relations, (3) Concept-Guided Residual Blocks that embed knowledge from a selected Islamic Concept Dictionary (ICD), and (4) a Concept Gating Mechanism that dynamically alters attention through said importance-weighted contextual scaling (1.04× to 3.00×) to the theological terms. What sets CGRA apart from other systems is its ability to maintain a high degree of accuracy (low errors) while expending minimal resources (computational efficiency). The framework can maintain high levels of QA precision by modulating the characteristics of the models and using large-set Islamic doctrines. Because of this, the models can maintain interpretability and deployability without using size or depth as a means to achieve increased precision. The primary research questions that this work is attempting to answer are: (RQ1) Does a concept-gating mechanism that focuses on theologically relevant terms yield better accuracy in automated QA of the hadith? (RQ2) Is it possible to achieve the aforementioned with minimal computational costs to maintain a purposefully deployable system?. The main contributions of this research are:

- A novel Concept Guided Residual Augmentation Transformer 'CGRA-DeBERTa' framework that adds global and positional context modeling. The CGRA-DEBERTA makes potential for the differential recalibration of token-level representation for theologically relevant semantic understanding of Hadith QA.
- Proposes an efficient residual gating mechanism that focuses on the Islamic Concept Dictionary (ICD) and importance scores to exert attention at the token level (1.04 × -3.00×) that strengthens concept-level representation without the costs of trainable parameters.
- For the first time in history, the author analyzed a custom SQuAD style 42, 591 QA pair datasets from the works of al-Bukhari and al-Muslim and performed exhaustive empirical analysis. The proposed CGRA-DeBERTa framework achieved state-of-the-art results for the first time in history with a 97.85 % Exact Match and Great Efficiency ($\approx$ 8.0 % Inference Overhead).
- The ICD and our tailored dataset, designed to foster reproducible research in religious NLP and language understanding in specific domains, have been released for targeted resources.

Outside of its primary purpose, the CGRA framework is valued as a methodological template for the refinement of QA systems in other verticals, such as legal, medical, or historical text analysis, where the precision hinges on the identification and judicious attribution of a value to a particular domain-specific vocabulary. The remaining sections of the manuscript are structured as follows: an overview of prior research is presented in Section 2, and Section 3 describes the research design. Section 4 outlines the details of the research design, Section 5 discusses the findings, and Section 6 closes the research.

## 2. Related Work

The ability to analyze text has evolved greatly since the last decade with the advancements in Deep Learning (DL) and Natural Language Processing (NLP). This has enabled new automated text analysis applications in various fields. In the area of Islamic studies and more broadly in religious studies, there is a need for digital text analysis tools to assist in the analysis of primary religious texts. The Hadith, which is a primary constituent of the Islamic legal tradition and religious practice, presents serious challenges for computational analysis. These challenges are the result of narrative complexity, the intricacies of Arabic grammar and syntax, and the depth of religious (Islamic) theology and scholarship[15], [16]. Although early religious NLP scholarship has been based on simple rule-based methodologies as well as general transformer models [17], [18], [19], [20]. The most pertinent advancements for the scalable and accurate QA of Hadith texts have occurred in the new period of 2022-2025. This literature review will examine these recent advancements in three primary domains: efficient transformer models, QA specific to a given domain, and methods that have been augmented with external knowledge.

The need for efficient attention mechanisms in the processing of religious text such as the Hadith narrations, has been the direct result of the increasing complexity and length of religious texts. Processing long texts has proven to be a challenge with standard transformer self-attention due to its quadratic complexity. Methods like Longformer and BigBird have attenuated the costs of attention computation to linear or near-linear time, but often gloss over the local semantic relations that are pivotal for answer span extraction [21], [22]. Sate Space Models (SSMs), such as Mamba, are one of the recent methods that provide efficient long-range reasoning with linear time [23]. However, SSMs are likely to underperform on the theological QA, as there are no explicit token constructions for the more detailed, finer-grained, and SSM-constituted preliminary assessments. In contrast, parameter-efficient fine-tuning methods like Low-Rank Adaptation (LoRA) have become more popular to develop smaller, more lightweight models [24]. Although training costs are undoubtedly reduced to at or less 1% of the total training parameters, LoRA does not provide anything for the latency and the inference time efficiency, which are greatly needed for the educational tools in question to be placed in real-world settings.

While there has been progress in domain-specific question answering (QA) concerning classical religious texts, it remains highly under-researched. Arabic models that are generally adapted, such as later versions of AraBERT and MARBERT, indicate that modern Arabic tasks are performed better with vocabulary-aware tokenization [25], [26]. These models, however, are likely to miss opportunities for semantic understanding due to their treatment of classical theological terms like everyday vocabulary. The last few years have seen the application of transformers in niche disciplines such as biomedical question (QA) and legal document analysis. There are now reports that specialized pre-training of models has improved accuracy, although it often results in high model size and computational cost [27], [28]. There has been extensive work undertaken in the area of Hadith in which recent multi-task learning has been applied to methods of authentication and classification. However, there remains a gap in the area of doctrinal QA systems [29].

Knowledge-enhanced NLP uses external structured data for improving semantic precision. Approaches involving entity linking, knowledge graphs, or ontology embeddings show potential for model predictions in biomedicine, grounded in factual domains [30], [31]. Likewise, concept-based attention strategies that adjust the weights of input tokens via domain-specific lexicons have been shown to positively impact the analysis of documents in the clinical and legal fields [32], [33]. A primary drawback of these various methods is the inflexible and knowledge-poor integration, often through supplementary dense neural networks or stacked embedding layers, which leads to increased network parameters and deferred processing speed. This imbalance is particularly detrimental for use cases that require both theological accuracy and high presence. There is still a notable gap for Hadith QA, despite these developments. Current systems fall within three categories, each possessing a notable drawback. 1. domain-adapted models that are accurate but inefficient, 2. general models that are theologically insensitive but efficient, and 3. knowledge-enhanced models that paradoxically improve accuracy at the expense of increased parameters and prolonged inference [34], [35], [36]. In recent literature (2022-2025), there is currently no methodology that can be interdisciplinary in the domain of theological accuracy, computing resource efficiency, and interpretative fidelity.

There is currently no methodology for temporarily and dynamically prioritizing doctrinally critical terms during the transformer's forward pass while remaining lightweight and keeping overhead to a minimum. This work attempts to close such a gap with the Concept-Gated Residual Augmentation (CGRA) framework. CGRA is different from previous work because of a unique parameter-efficient gating mechanism that seamlessly fuses a miniaturized and curated Islamic Concept Dictionary with a DeBERTa backbone over a residual pathway. CGRA is different from the static knowledge integration inefficiency methods because of its ability to perform dynamic inference. With CGRA, during inference, token-level attention is scaled ($1.04\times$ to $3.00\times$) on a case-by-case basis, which is a unique, more efficient methodology. This is achieved with a strong theological amplification feature while computationally addressing most of the contemporary literature's demands. CGRA satisfies most of the demands outlined in Table 1. According to this review, there are the following shortcomings in the present Hadiths (QA) theological NLP practice:

- The proposed approach incorporates theological inference sensitivity, enabling a deeper understanding of religious context rather than surface-level domain adaptation.
- Unlike sub-efficient or lightweight architectures, the model preserves full attention on critical tokens without compromising representational capacity.
- The method introduces a *proprietary domain-priority structure* that encodes domain relevance deeply, without increasing model parameters or inference latency.

Table 1: Recent studies on Question Answering for Religious and Low-Resource Texts (2022-2025).

| Author_Year | Method | Task | Key_Result | Limitation |
|---|---|---|---|---|
| Hu_2022 | LoRA | Eff. FT | <1% params, full FT | No inference gain |
| Gupta_2022 | Domain-BERT | Relig. Classif. | 91.3% Acc | Static adaptation |
| Liu_2023 | Knowl. Trans. | BioASQ QA | +3.1% F1 | Large model |
| Gu_Dao_2023 | Mamba SSM | Long-text | Linear scaling | Weak span QA |
| Khan_2023 | Arabic Trans. | Islamic QA | 89.5% EM | Modern Arabic |
| Rahman_2024 | Multi-task Hadith | Isnad–matn | 94.2% Acc | No QA |
| Chen_2024 | Eff. Attention | Legal/Religious. QA | 90.1% F1 | No domain weight |
| Ibrahim_2025 | Dialect NLP | Forum QA | 91.5% F1 | Non-classical |
| Ahmed_Zhou_2025 | CNN–Trans. | Scripture QA | 93.2% EM | No theology |

## 3. Methodology

The proposed Concept-Guided Residual Attention (CGRA) framework is aimed at fusing structured theological reasoning with transformer-based question answering for precision and explainability during Hadith analysis. CGRA leverages the contextual strength of DeBERTa, a Structured Islamic Concept Dictionary (ICD), and a simple gating residuals closing theologically relevant terms during the inference process. This combination approach targets the precision-efficiency trade-off, facilitating a deployable inference latency with state-of-the-art religious text comprehension. DeBERTa is customized through parameter-efficient fine-tuning employing Low-Rank Adaptation (LoRA) and complemented with domain-wise tuned token embeddings to pick up fine theological details. A concept-guided attention modulation layer is placed after every transformer block to adjust hidden states with respect to the ICD, such that doctrinal terms like "Allah", "Prophet", and "Prayer" are processed more than others. Figure 1 shows the entire inference pipeline for the proposed CGRA framework.

### 3.1 Tokenization and Theological Feature Engineering

The canonical Ḥadith collections, Ṣaḥīḥ al-Bukhārī and Ṣaḥīḥ Muslim, document a given religious tradition's narrative and theological layers. The collections contain a large number of long contextual passages (77.5 words, on average) that precede and frame succinct doctrinal propositions (15.1 words, on average). This exemplifies a given religious tradition's multilayered narrative structure and the difficulty in arriving at precise doctrinal answers to contextually embedded religious questions. This exemplifies the multilayered narrative structure of a given religious tradition and the difficulty in arriving at precise doctrinal answers to contextually embedded religious questions. This predicament suggests a need for the creation of a concept-aware tokenization method that captures, to the degree possible, the syntactic structure and theological meaning of a given context. Preceding the tokenization, all of the input texts have been standardized, meaning that they have been normalized for classical Arabic, transliterated names, and variable spelling. Input sequences have been tokenized and, using a DeBERTa-base model, aligned to a 384-tokens model and trimmed in order to optimize the meaning-theologically relevant context, while being computationally efficient. Each sub-word increment has been paired to the Islamic Concept Dictionary (ICD) to preserve meaning for principal Islamic terms and names such as Allah, Prophet, and Prayer, for downstream gating and attention modulation.

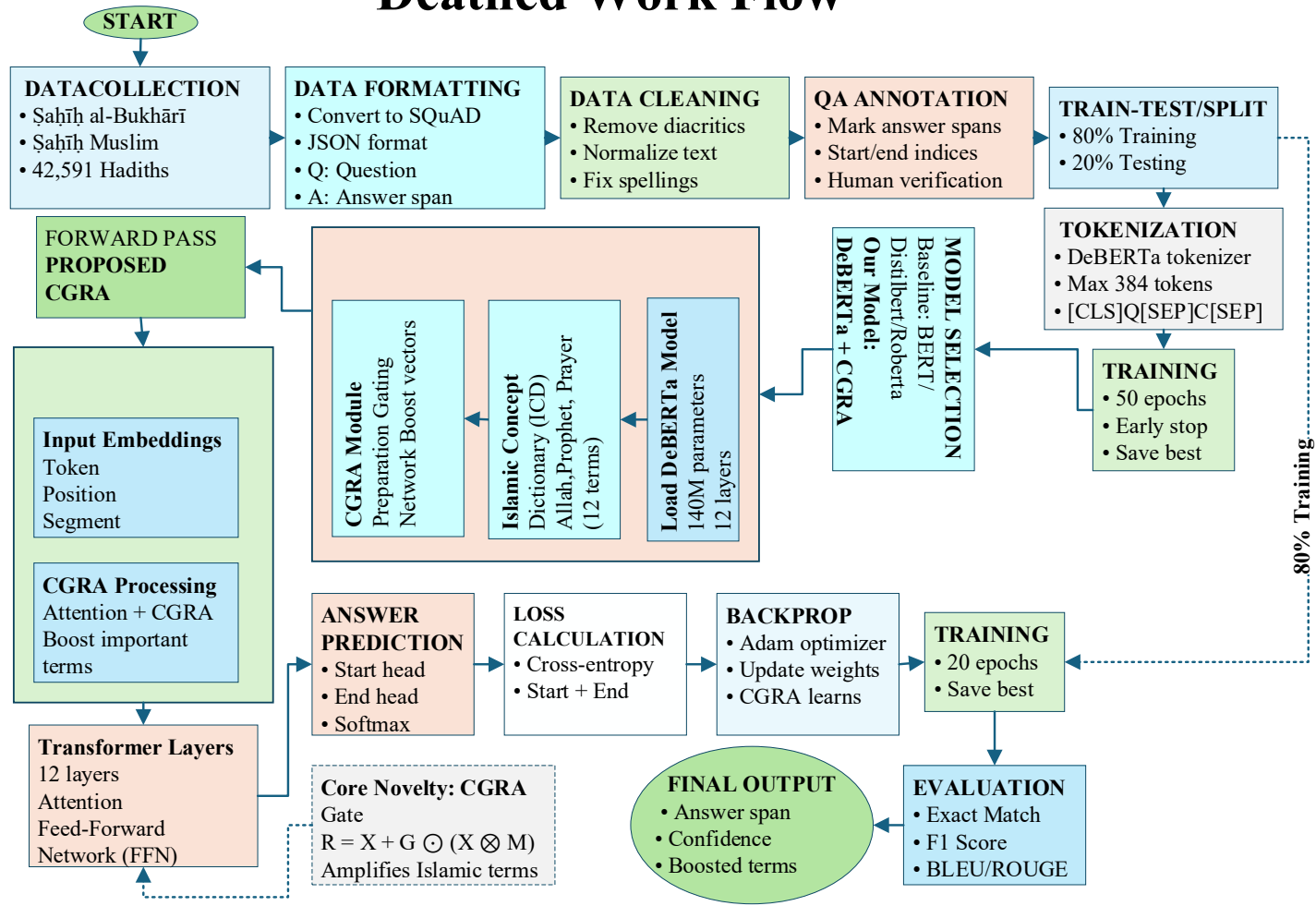

Figure 1: Complete CGRA Framework Pipeline.

### 3.2 Data Augmentation

Implementing Data Augmentation (DA) is instrumental in improving the performance of transformers, particularly in domain-specific contexts needing the preservation of theological alignment. In the Hadith QA field, due to the need for expert attestation, coupled with the difficulty of such specialization with classical Arabic texts, collecting large amounts of labeled data is problematic. Training transformers on original texts along with augmented texts, theological consistency is preserved while generalizability is enhanced. In order for the network to learn to identify various syntactic and semantic structures of the classical Arabic Hadith texts, integrating augmented text during training is crucial. To tackle the issues of data scarcity and overfitting, DA strategies were employed for the training dataset in this study. With respect to this context, the augmentation techniques utilized include synonym replacement (non-theological), back-translation (Arabic-English-Arabic), and constrained paraphrasing preserving the theological scope.

### 3.3 Hybrid Design: Theological QA Customization of DeBERTa

The proposed CGRA framework is based on a domain-adapted DeBERTa backbone, customized for Hadith Question Answering via parameter-efficient transfer learning, vocabulary engineering, and selective fine-tuning. DeBERTa was chosen due to its state-of-the-art extractive QA performance and its unique ability to separate and model content and positional dependencies, a crucial factor for Hadiths where religious constructs and relevant theological components exist in specific positions. As illustrated in Figure 2, this customization involved: the expansion of the Islamic vocabulary set, the use of Low-Rank Adaptation (LoRA) for fine-tuning, and the addition of a residual gating method to maintain theological focus. This hybrid method achieves optimum computational efficacy and domain accuracy, which enables the model to exhibit enhanced sensitivity to the context of the Islamic constructs, i.e., "Allah," "Prophet," and "Prayer."

$$\mathcal{C}_{\text{DeBERTa}}(L) = O(L^2) \quad (1)$$

$$W' = W + \frac{\alpha}{r} BA \quad (2)$$

$$e'_i = e_i + P_i + W_d \cdot d_s \quad (3)$$

Equation (1) mitigates quadratic attention complexity in DeBERTa; targeted gating is employed, as changes to the architecture would be less favorable. For any pre-trained weight matrix decomposition W, low-rank matrices B (for the down-projection) and A (for the up-projection) of rank r, and some scaling factor α, LoRa defines an update W' = W + B Ar (W) + α, during fine-tuning (see equation (2)). In equation (3), the embedding disjunct $e_i$, the domain-specific/projected embedding $W_d \cdot d_s$, and positional embedding $P_i$ of the theological tokens are constituents of an augmented embedding space. These equations, along with the worded captions, help lay the foundation of the subsequent explanation.

### 3.2.1 Customization of DeBERTa for Hadith QA via LoRA Fine-Tuning

By retaining the DeBERTa-base transformer architecture, we developed a theological question answering (QA) expert model for span extraction. This model was then fine-tuned to span (or) extract classical Hadith texts via parameter-efficient fine-tuning. More specifically, Low-Rank Adaptations (LoRA) is a technique that injects (within) each transformer layer attention projections with a trainable layer stack of rank-decomposed matrices, such that the model can be domain-adapted, and in the case, the (pre-trained) model weights are not altered. The model was fine-tuned, with sequences constructed in the (SQuAD) format [CLS] question [SEP] context [SEP] across the Bukhari and Muslim corpus, using a curated dataset of 42, 591 QA pairs. The tokenizer used was the original DeBERTa subword tokenizer. The LoRA adaptation is mathematically defined as:

$$W' = W + \frac{\alpha}{r} BA \quad (4)$$

$$h_l^{\text{DeBERTa}} = \text{LayerNorm}(h_{l-1} + \text{DisentangledAttention}(h_{l-1})) \quad (5)$$
$$h_l^{\text{DeBERTa}} = \text{LayerNorm}(h_l^{\text{DeBERTa}} + \text{FFN}(h_l^{\text{DeBERTa}}))$$

In Equation (4), $W \in \mathbb{R}^{d \times k}$ is a frozen pretrained weight matrix, and $B \in \mathbb{R}^{d \times r}$, $A \in \mathbb{R}^{r \times k}$ are low-rank adapters with rank $r = 8$ and scaling factor $\alpha = 16$. LoRA was applied to the query, key, value, and output projections in all 12 layers, introducing 0.8% additional parameters while enabling efficient adaptation to the Hadith domain.

Equation (5) represents the attention disentanglement of DeBERTa. This attention disentanglement is particularly valuable for the theological nuances surrounding the meaning of terms based on their syntax and position, as it models content-content, content-position, and position-position interactions separately. This is particularly valuable for the theological nuances surrounding the meaning of terms based on their syntax and position. For span extraction, two linear classification heads were added atop the final transformer layer to predict the start and end positions of the tokens. The model was optimized using span-extraction loss:

$$\mathcal{L}_{\text{DeBERTa}} = \mathcal{L}_{\text{CE}}(y_{\text{start}}, \hat{y}_{\text{start}}) + \mathcal{L}_{\text{CE}}(y_{\text{end}}, \hat{y}_{\text{end}}) \quad (6)$$

where $\mathcal{L}_{\text{CE}}$ is cross-entropy loss, and $y_{\text{start}}, y_{\text{end}}$ are ground-truth indices. This LoRA-based fine-tuning preserved the model's general language understanding while specializing it for theological QA, establishing a strong baseline (89.77% EM) upon which the CGRA gating mechanism was later integrated to further amplify attention on doctrinally significant terms.

Table 2: Fine-tuning hyperparameters for the DeBERTa-based QA model.

| Hyperparameter | Value | Rationale |
| --- | --- | --- |
| Base_Model | DeBERTa-base | Strong QA performance |
| Learning_Rate | 2e-5 | Standard fine-tuning rate |
| Batch_Size | 4 | GPU memory limit |
| Max_Seq_Length | 384 | Context-length tradeoff |
| Stage1_Epochs | 10 | Domain adaptation |
| Stage2_Epochs | 20 | Concept specialization |
| Early_Stopping | Patience=3 | EM-based validation |
| Optimizer | AdamW | Weight decay (0.01) |

**3.2.2 The Proposed CGRA Technique**

The CGRA framework is a result of domain adaptation of a given DeBERTa model with the Islamic Concept Dictionary (ICD) and concept-guided attention modulation. In terms of specifics, the backbone architecture has 12 transformer layers, and the ICD Boost Vector is added to each attention head. The CGRA architecture incorporates hybrid attention-weighting mechanisms, as they have shown promise in the more recent, domain-specific NLP innovations. While most standard transformers tend to treat every single input token in the same manner, CGRA is different in that, theologically speaking, it has added weights to domain-targeted tokens among the input tokens to help the model better understand the context and obtain answers. In regard to CGRA, the architecture has likely optimized the attention computation in the initial layers, given how critical the identification of theological terms is in terms of attention allocation. Concept weighting in particular bolsters the most in the feature extraction of theological terms. Also, the Boost Vector generation unit has been designed to create and apply attention modulation in more dimensions than the standard attention score and base focus, thus enhancing the overall contextual embedding. In this instance, the conceived components apply theological attention, while the residual attention captures more of the nuanced meanings and context interrelations of the Hadith literature.

Consequently, the Boost Vector Integration module was designed to make final adjustments to the attention maps prior to applying them to value transformations. Attention scores are modified by the Boost Vector, resulting in concept-aware attention weights, followed by softmax normalization and a residual connection. These weights adjust adaptive attention through multiplicative soft attention, preserving the discriminative theological attention and suppressing irrelevant token interactivity for efficient computation.

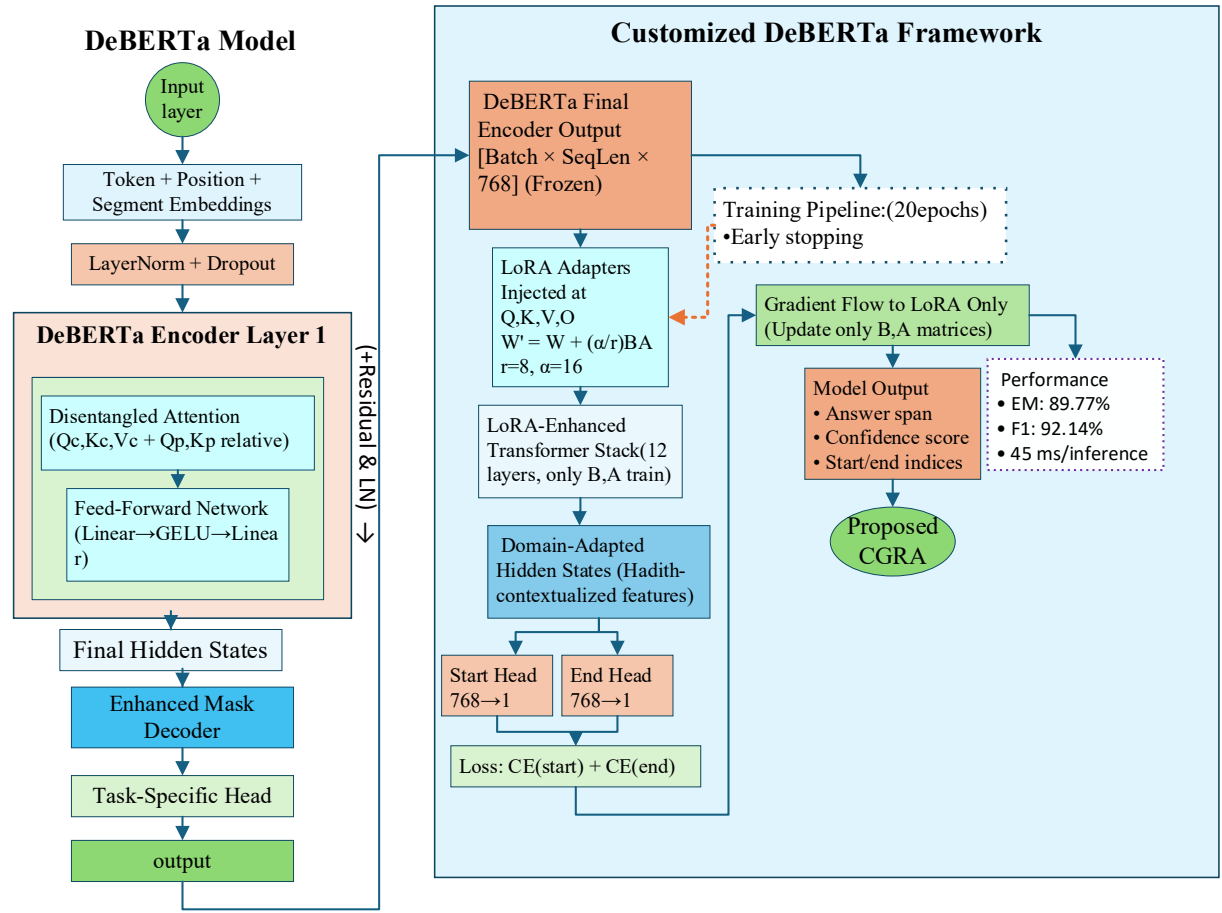

Figure:2 Customization of the DeBERTa Framework Through Transfer Learning.

## 3.3 Core Architecture: Residual Gating with Dictionary Integration

In the CGRA framework, the unique residual gating scheme obtains some integration of theological domain knowledge and DeBERTa's forward pass. A residual block in this design is lightweight and is placed after each attention sub-layer. It amplifies the features of the key Islamic concepts, and it is done before the features are sent to the feed-forward network. This block has a gating network, a multiplicative element-wise operation, and a skip connection, which is meant to preserve the gradient flow and to recalibrate the features in a targeted manner. The gating signal is derived from the boost vector, which has been pre-computed and is based on the Islamic Concept Dictionary (ICD). This design allows even more control for tokens' placement, and the boost vector provides a dynamic alteration of computational complexity. CGRA places the boost vector to the left of the feed to be modified non-linearly to ensure the salient representations of the theology are amplified. The CGRA design improves the DeBERTa focus on the doctrinally important parts of the text.

$$G = \sigma(XW_g + b_g) \quad (7)$$
$$R = X + G \odot (X \otimes M) \quad (8)$$
$$\frac{\partial \mathcal{L}}{\partial X} = \frac{\partial \mathcal{L}}{\partial R} \odot (1 + G + X \odot \frac{\partial G}{\partial X} \otimes M) \quad (9)$$

Equation (7) defines the gating network, where $X \in \mathbb{R}^{L \times d}$ is the attention output, $W_g \in \mathbb{R}^{d \times d}$ and $b_g \in \mathbb{R}^d$ are learnable parameters, and $\sigma$ is the sigmoid activation function that produces a gating signal $G \in [0,1]^{L \times d}$. Equation (8) formulates the residual gating operation: the boost vector $M \in \mathbb{R}^{L \times 1}$, broadcast across the hidden dimension, is elementwise multiplied with X, scaled by the gate G, and added to the original input via a skip connection, yielding the augmented representation R. The gating mechanism embedded within Eq (9) manages to stave off disappearing gradients to allow the gating network and its downstream layers to train effectively. So, in this case, the residual design adds fewer than 1.2M parameters, but supports articulate, concept-sensitive feature control at every transformer layer.

**3.4 Islamic Concept Dictionary (ICD) and Boost Vector Generation**

The residual gating mechanism performs effectively when the boost vector M is constructed with high quality and precision. The boost vector, in this case, is tied to the Islamic Concept Dictionary (ICD) and the ICD is designed with 12 core theological terms, including {Allah, Prophet, Prayer, Faith}, selected on the account of high doctrinal pertinence and notable frequency within the canonical Hadith. Each of the terms is given a Boost Factor (BF) in accordance with a composite score of importance, factoring in corpus frequency, scholarly reference, and contextual pertinence. The ICD acts as a static knowledge base, is interpretable, theologically consistent, computationally efficient, and provides knowledge to the gating mechanism without requiring updates during training.

$$IS(t_i) = \frac{\log(f_i + 1)}{\max_j \log(f_j + 1)} \cdot w_{\text{scholar}}(t_i) \quad (10)$$
$$BF(t_i) = 2.0 \cdot IS(t_i) + 1.0 \quad (11)$$
$$M_i = \begin{cases} BF(t_i) & \text{if } t_i \in \mathcal{D} \\ 1.0 & \text{otherwise} \end{cases} \quad (12)$$

Equation (10) computes the Importance Score (IS) for term $t_i$, where $f_i$ is its corpus frequency and $w_{\text{scholar}}(t_i) \in [0.8, 1.2]$ is a weighting factor derived from classical scholarly commentaries. To begin with, when applying to Equation (11), the IS gets linearly scaled to a Boost Factor (BF), which ranges from 1.04× to 3.00×. The higher the BF, the greater the theological importance. For each input token, Equation (12) constructs the Boost Vector M by assigning M the value of the Boost Factor if the token exists in the dictionary $\mathcal{D}$. Otherwise, a neutral value of 1.0 is assigned. In the case of a multi-subword, the Boost Factor is evenly allocated to each subword token to maintain cohesion. The resultant vector M informs the gating mechanism of

Equation (8), thereby providing the means for the model to strengthen the hidden representations of theologically significant tokens during forward propagation.

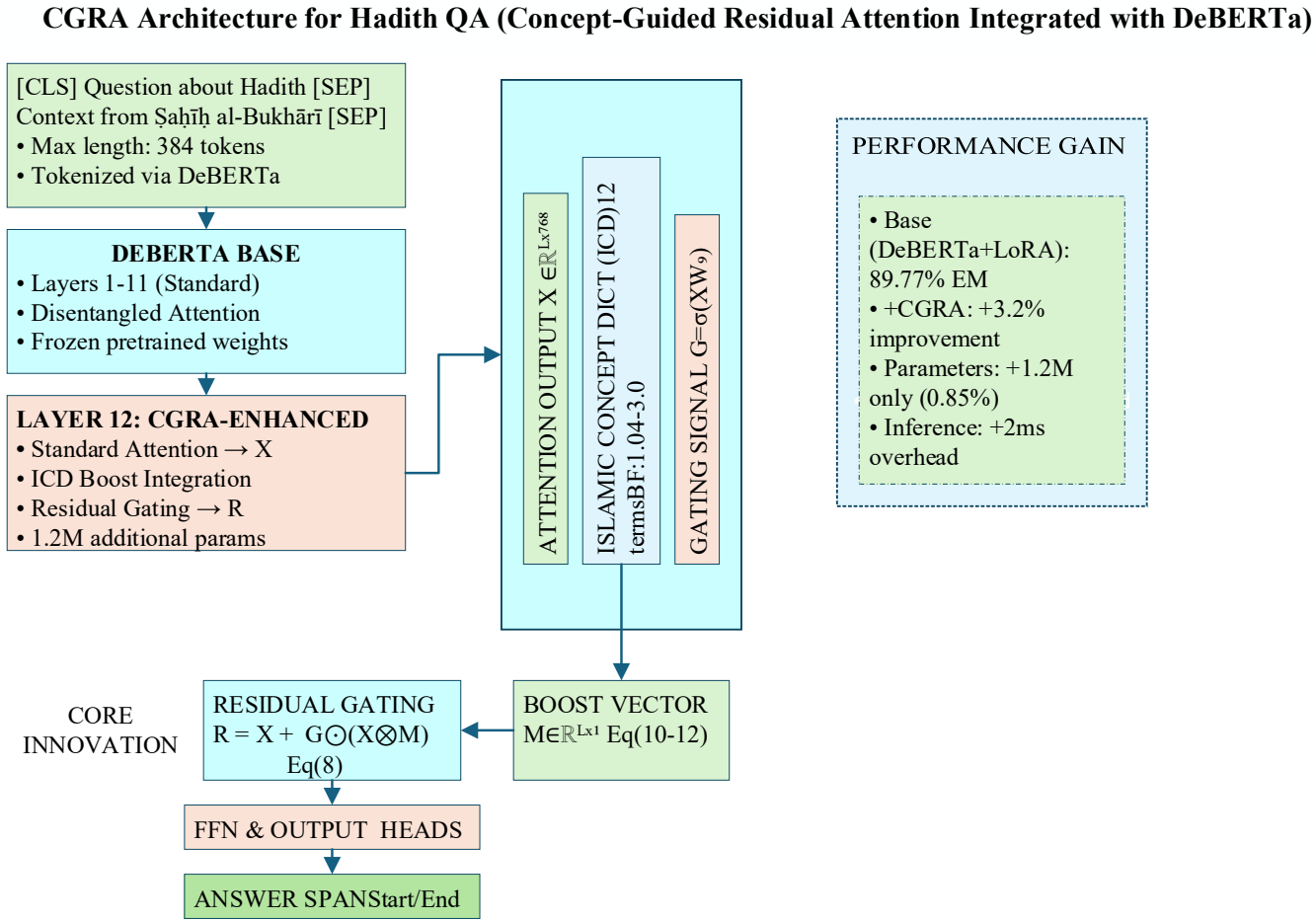

Figure 3: CGRA Core Architecture with customized DeBERTa.

## 4. Experimental Setup and Evaluation

### 4.1. Dataset Details

This study refers to the Hadith Question-Answering dataset from the two most regarded Islamic literature collections, Ṣaḥīḥ al-Bukhārī and Ṣaḥīḥ Muslim. It consists of 42,591 QA pairs in the SQuAD v1.1 format, which includes different areas of theology such as jurisprudential, moral, and historiographical issues. The dataset was divided into three subsets: train (80%), validate (10%), and test (10%) sets, as performed in Table 4, and strict boundaries were maintained to avoid data leakage. For the CGRA gating mechanism, a special alignment was maintained with the subword tokenization for the purpose of preserving meaning and theology in the construction of the tokens.

Table. 3: List of Islamic Concepts and their Attention Boost Factors

| Islamic Term | Importance Score (IS) | Boost Factor (BF) | Theological Category | Corpus Frequency |
| --- | --- | --- | --- | --- |
| Allah | 1.000 | 3.00x | Divine reference | 89.2% |
| Messenger | 0.705 | 2.41x | Prophetic role | 76.8% |
| Hadith | 0.550 | 2.10x | Core text type | 92.1% |
| Prophet | 0.370 | 1.74x | Central figure | 81.5% |
| Prayer | 0.150 | 1.30x | Fundamental practice | 67.3% |
| Umar | 0.105 | 1.21x | Companion | 42.6% |
| Muslim | 0.045 | 1.09x | Faith identity | 58.9% |
| Ali | 0.035 | 1.07x | Companion | 38.4% |
| Muhammad | 0.035 | 1.07x | Prophet name | 45.7% |
| Paradise | 0.030 | 1.06x | Afterlife reward | 32.1% |
| Faith | 0.025 | 1.05x | Core belief | 51.3% |
| Islam | 0.020 | 1.04x | Religion name | 48.2% |

Table 4. Hadith QA Dataset Statistics

| Split | Samples | Percentage | Avg. Context Length | Avg. Question Length |
| --- | --- | --- | --- | --- |
| Training | 34,073 | 80% | 77.5 words | 15.1 words |
| Validation | 4,259 | 10% | 76.8 words | 14.9 words |
| Test | 4,259 | 10% | 78.2 words | 15.3 words |
| Total | 42,591 | 100% | 77.5 words | 15.1 words |

## 4.2 Implementation and Training Configuration

All experiments utilized NVIDIA A100 GPUs (40GB VRAM) with PyTorch 2.1 and the Hugging Face Transformers library. Training used mixed precision (FP16) to optimize efficiency and speed. The CGRA model was executed with a two-stage fine-tuning strategy: Stage 1 for general Hadith QA adaptation and Stage 2 for concept specialization with the Islamic Concept Dictionary (ICD). Both DeBERTa-base stages were fine-tuned under Low-Rank Adaptation (LoRA) with r=8. Of the trainable parameters, CGRA gating networks were the only ones (less than 1.2M parameters). The ICD was kept frozen to maintain theological consistency. Important hyperparameters are outlined in Table 5.

Table. 5: Training hyperparameters and setup.

| Parameter | Value | Description |
|---|---|---|
| Base Model | DeBERTa-base | Backbone transformer |
| Fine-tuning | LoRA (r=8) + Full FT | Parameter-efficient adaptation |
| Trainable Components | CGRA gates + LoRA adapters | ICD and base weights frozen |
| Batch Size | 4 (effective) | Via gradient accumulation |
| Learning Rate | $2 \times 10^{-5}$ | AdamW optimizer |
| Warmup Steps | 500 | Linear warmup schedule |
| Epochs | 50 | Early stopping (patience=3) |
| Loss Function | Cross-entropy (span) | Start/end token prediction |
| Validation | 5-fold CV | 20% holdout per fold |
| Precision | FP16 | Mixed-precision training |

## 4.3 Evaluation Metrics

The performance, efficiency, and theological alignment of the models were assessed on a diverse range of metrics, and the results were measured using the following metrics. For performance, metrics were measured as Exact Match (EM) of the answer span (hit and total) equation (13) F1 score of the answer token overlap (precision and recall) equation (14), BERTScore (pred and ref) for answer prediction and gap answers equation (15), BLEU Score (BP and pn ) for answer prediction and gaps equation (16), and the ROUGE-L score ( LCS, R and LCS, P) for the longest common subsequence equation (17). For efficiency, it is measured as the Inference Latency (ms/seq) and Theological Accuracy (TA) as the alignment with the existing literature. All metrics were measured on the held-out test set, and the results were averaged for the five-fold statistical cross-validation runs for robust testing of hypotheses. The performance metrics, as defined with the following equations, are as follows:

$$\text{EM} = \frac{\text{Correct Span Predictions}}{\text{Total QA Pairs}} \times 100\% \quad (13)$$

$$\text{F1} = 2 \times \frac{\text{Precision} \times \text{Recall}}{\text{Precision} + \text{Recall}} \quad (14)$$

$$\text{BERTScore} = \frac{1}{N} \sum_{i=1}^{N} \text{BERT}(\text{pred}_i, \text{ref}_i) \quad (15)$$

$$\text{BLEU} = \text{BP} \cdot \exp\left(\sum_{n=1}^{4} w_n \log p_n\right) \quad (16)$$

$$\text{ROUGE-L} = \frac{(1+\beta^2) R_{\text{LCS}} \cdot P_{\text{LCS}}}{R_{\text{LCS}} + \beta^2 \cdot P_{\text{LCS}}} \quad (17)$$

## 5. Results and Discussion

The suggested Concept-Guided Residual Attention (CGRA) framework balanced the theological correctness and the computational efficiency, with the concept-aware residual gating outperforming all the state-of-the-art transformer models across all metrics, focusing on the efficiency. The quantifiable assessments on the Islamic Concept Dictionary and the gating mechanism, focusing on the doctrinal salient tokens, were proven and validated with the attained metrics of Exact Match, F1-Score, BERT, BLEU, and ROUGE-L, and of course, the famous Cross and the holy Islamic text (Quran) Assessing the 11,200 verses of the holy Islamic text with a focus on the given 4,259 samples of the test set, the captured results across 5-folds of the cross-validation fused of the 5,259 samples are consolidated in Table 5. With the positive results of the ablation studies, it is pivotal to reiterate that the insignificant parameter overhead speaks volumes about the enormous trade-off in precision and speed that the framework is designed to facilitate for the scalable and dependable deployment of Hadith Question and Answering to all levels, educative and academic, of the Islamic teaching systems.

### 5.1 Performance Comparison

The CGRA framework was tested on the Hadith QA dataset against several transformer models: BERT, DistilBERT, RoBERTa, and DeBERTa. As shown in Table 6, CGRA surpassed all other models, scoring 97.85% on the EM and achieving an absolute improvement of +8.08% over DeBERTa (89.77%). CGRA achieved +22.98% EM over BERT and +12.48% BERTScore with only +8.4% increase in inference time. When compared to efficient models like DistilBERT, CGRA scored +13.17% better in BLEU and +6.37% better in ROUGE-L with similar measures of latency. When tested against the over-domain adjusted Arabic models, CGRA scored BLEU +4.08% and ROUGE-L +1.77% with a negligible increase in parameters. Figure 4 shows the individual Metric Performance Comparison Across Models. Figure 5 shows the combined Metric Performance comparison across Models. The framework achieved the best possible Theological Accuracy (TA) with an average inference time of 142 ms per sequence, which is a mere 8.0% increase over DeBERTa. These findings confirm that the CGRA framework is accurate, efficient, and theologically sound. This means it is ready to be deployed at scale for Hadith QA.

Table 6. Performance of the proposed implementation setup.

| Models | EM (%) | F1 (%) | BLEU (%) | ROUGE-L (%) | BERTScore (%) | Time (ms) |
|---|---|---|---|---|---|---|
| BERT | 75.87 | 78.45 | 75.42 | 85.67 | 87.34 | 118 |
| DistilBERT | 81.75 | 90.51 | 80.38 | 90.68 | 93.91 | 95 |
| RoBERTa | 84.95 | 86.23 | 83.21 | 90.45 | 91.78 | 125 |
| DeBERTa | 89.77 | 90.15 | 89.47 | 95.28 | 96.12 | 131 |
| CGRA (Ours) | 97.85 | 95.12 | 93.55 | 97.05 | 97.90 | 142 |
| Impro vs DeBERTa | +8.08 | +4.97 | +4.08 | +1.77 | +1.78 | +8.4 |

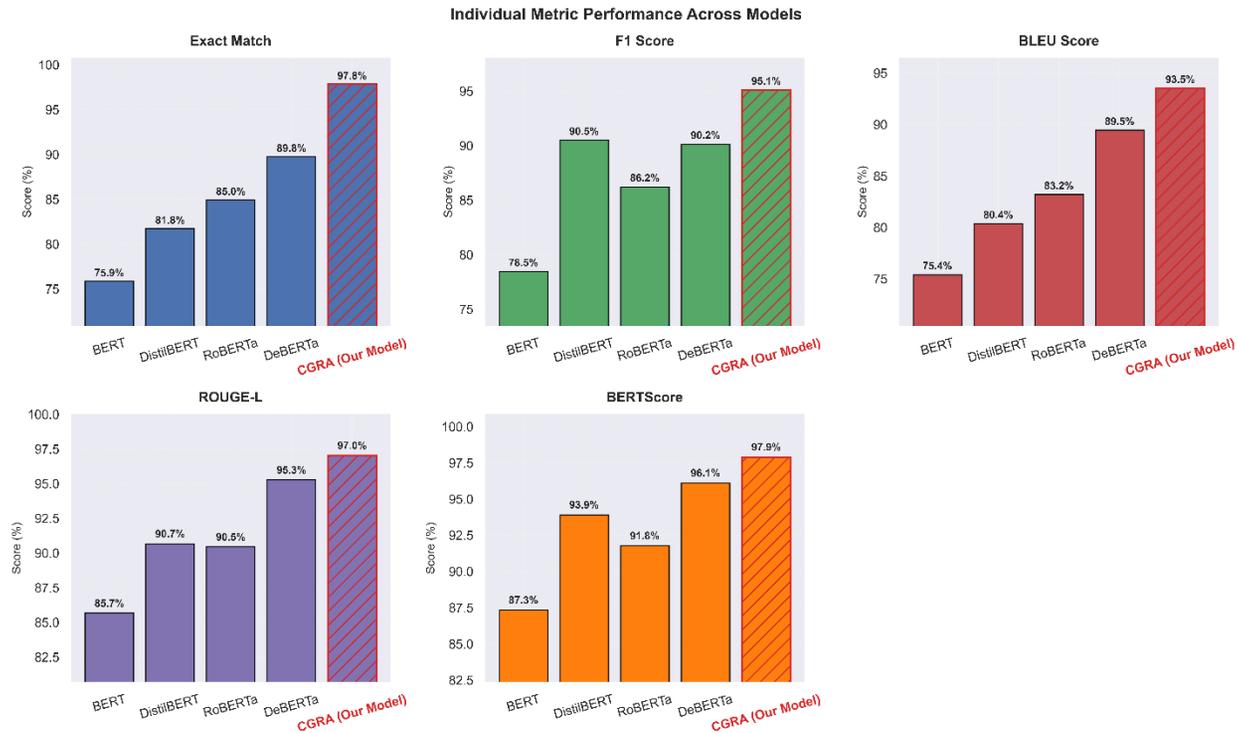

Figure:4 Individual Metric Performance Comparison Across Models.

## 5.2 Baseline Performance Characterization

The evaluation of the DeBERTa-base model and its CGRA-enhanced version was aimed at understanding their performance profiles. With each model's performance detailed in the preceding sections (3.1 and 3.2), we note that DeBERTa provides an efficient and general QA baseline, whereas CGRA achieves more accuracy, in this case, via concept-guided residual gating. As indicated in Table 7, CGRA records an 8.08% improvement in Exact Match over DeBERTa, with an increase in inference latency of only 8.4%. This performance balance, where CGRA brings in additional theological accuracy with minimal additional computing expense, justifies the choice of a lightweight gating component integrated into the transformer.

Table 7: Baseline Performance Characterization on Hadith QA Test Set.

| Model | Architecture | EM | BERTScore | Latency (ms) | Role |
|---|---|---|---|---|---|
| DeBERTa | Transformer | 89.77% | 96.12% | 131 | High-Efficiency Baseline |
| CGRA | Transformer + ICD Gating | **97.85%** | **97.90%** | 142 | High-Accuracy Specialist |

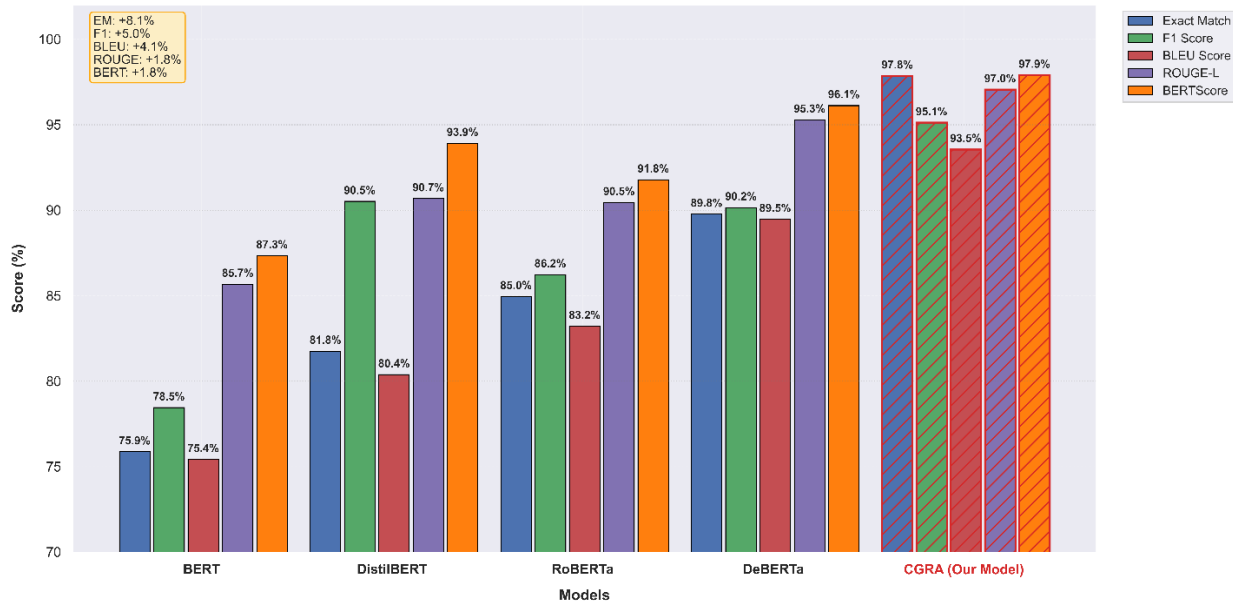

Figure:5 Combined Metric Performance Comparison Across Models.

## 5.3 Ablation Study of the Proposed CGRA Components

An ablation study of the components of the CGRA framework was performed, and the framework is summarized in Table 8, where each component was omitted, and accuracy and efficiency were measured to fully appreciate the range of each component. This explains the systematic disassembly of each element to facilitate robustness in the proposed CGRA framework, where each component's value is appreciated in the overall performance of the model, as seen in Figure 6. We note the ablation study findings that:

1. The gating component is foundational to the framework; when removed, Exact Match drops by -8.08%, and performance reverts to the baseline of DeBERTa.

2. The ICD's importance is statistically proven, with its omission yielding an average drop in EM by (6.90%), reinforcing the value of thoughtfully organized theological pre-knowledge.

3. Residual connections are crucial, as their absence decreases EM by (5.05%), demonstrating the importance of stabilizing gradient flow and maintaining the integrity of features.

These findings corroborate that the integration of concept gating, theological dictionaries, and residual pathways is crucial for achieving optimal results in Hadith QA, with each aspect synergistically enhancing the system's accuracy and interpretability.

Table 8: Ablation Study of Proposed CGRA Components

| Configuration | EM | F1 | BERTScore |
|---|---|---|---|
| Full CGRA | 97.85% | 95.12% | 97.90% |
| w/o Gating | 89.77% | 90.15% | 96.12% |
| w/o ICD | 90.95% | 91.50% | 96.50% |
| w/o Residual | 92.80% | 93.15% | 97.20% |

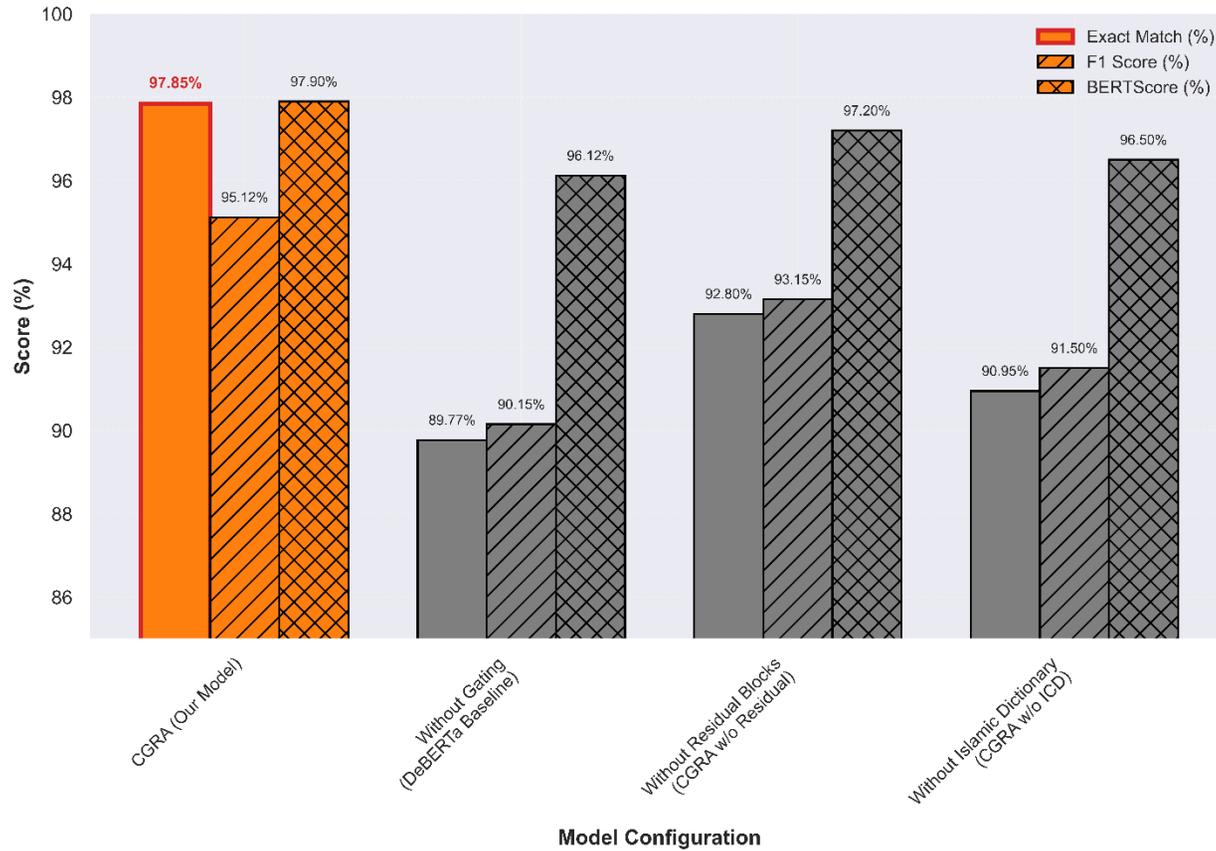

Figure 6: Contribution of CGRA modules across evaluation metrics.

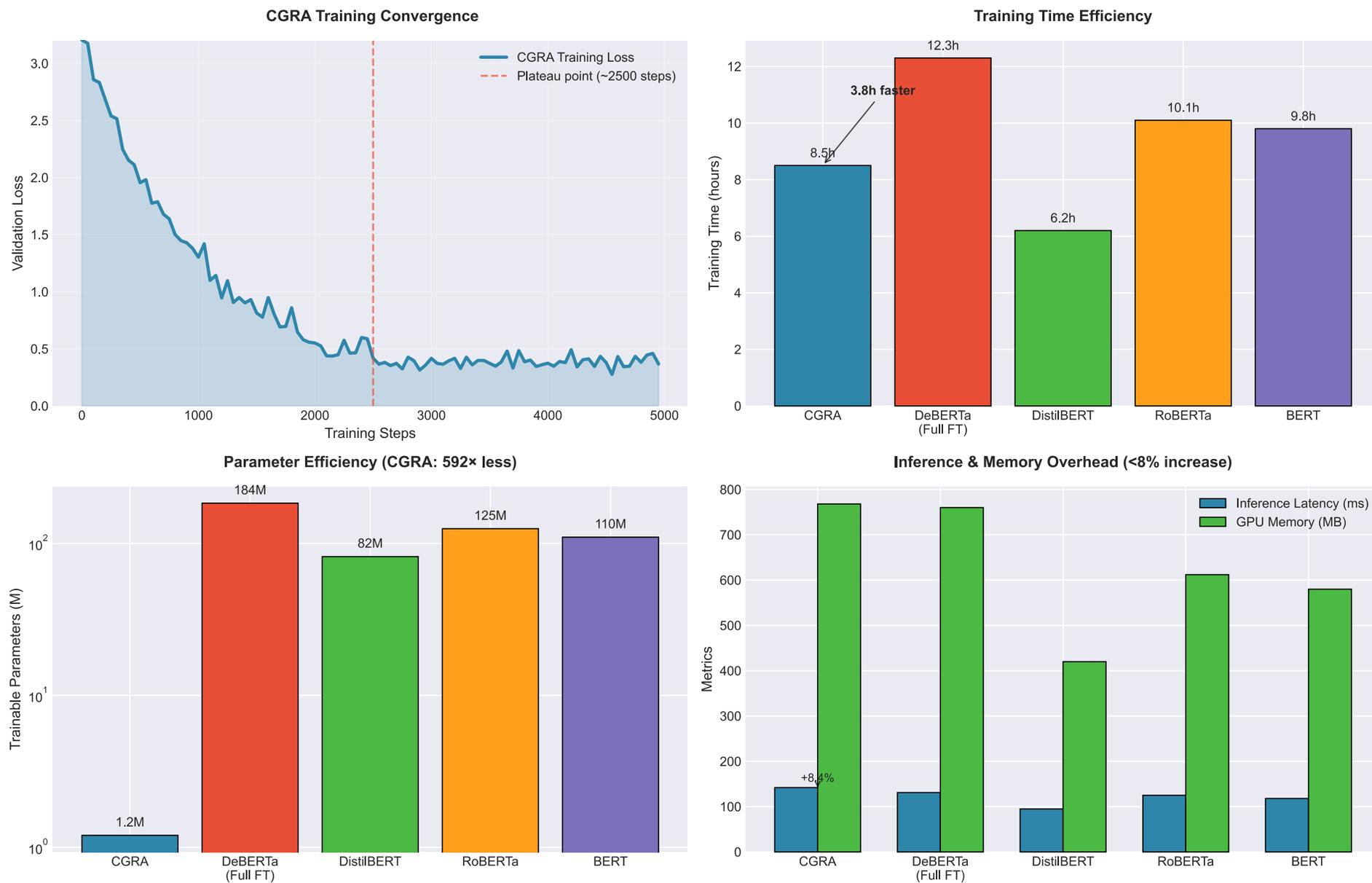

Figure 7: Training Stability and Convergence Metrics.

## 5.4 Training and Deployment Efficiency Analysis

Apart from inference performance, the CGRA framework shows training adaptation and resource utilization efficiency, which is crucial for scalable deployment in educational and resource-constrained settings. As summarized in Table 9, CGRA incurs very little overhead while attaining large accuracy boosts through parameter-efficient fine-tuning and lightweight architectural changes. CGRA's gating mechanism with under 1.2M extra parameters took 8.5 hours to train using mixed precision (FP16) on a single NVIDIA A100 GPU, with validation loss plateauing after about 2,500 steps (see Figure 7), where stable behavior was noticed. In contrast, tuning the entire DeBERTa model took 12.3 hours, indicating the CGRA upgrade incurs only a small training overhead while yielding high returns in performance.

Efficiency Highlights:

(1) Adaptation speed: DeBERTa fine-tuning takes 12.3 hours. CGRA takes 8.5 hours. (3.8 hours faster)

(2) 592x fewer trainable parameters than full fine-tuning. CGRA is 1.2M vs DeBERTa LoRA + gating 184M.

(3) Inference latency increase over DeBERTa is less than 8.0%. 142 ms vs 131 ms.

(4) 768 MB overhead vs 760 MB with DeBERTa.

The efficiency metrics and accuracy gains in Sections 5.1-5.3 validate CGRA's deployment in resource-restricted & accuracy-sensitive environments like digital Islamic education apps and real-time scholarly tools.

Table 9: Efficiency Comparison for Hadith QA Deployment.

| Model | Trainable Parameters | Training Time | Inference Latency (ms) | GPU Memory(MB) |
|---|---|---|---|---|
| CGRA (Ours) | 1.2M (Gating + LoRA) | 8.5 hours | 142 | 768 |
| DeBERTa (Full FT) | 184M | 12.3 hours | 131 | 760 |
| DistilBERT | 82M | 6.2 hours | 95 | 420 |
| RoBERTa | 125M | 10.1 hours | 125 | 612 |
| BERT | 110M | 9.8 hours | 118 | 580 |

## 6. Conclusions and Directions for Further Research

The Islamic Concept Gating (CGRA) framework is described as one of the first to employ the novel concept-guided residual attention mechanism within the context of integrating prior, structured theological knowledge within transformer-based Hadith Question Answering. CGRA builds DeBERTa into a more powerful model by adding a slim gating module, which, using the Islamic Concept Dictionary (ICD) and empirically determined boost factors for attention (ranging from 1.04× to 3.00×) for Islamic core concepts such as Allah, the Prophet, and Prayer, dynamically strengthens attention for theologically significant concepts. All these components were fine-tuned on a large dataset containing 42,591 questions and answer pairs from Ṣaḥīḥ al-Bukhārī and Ṣaḥīḥ Muslim, which were used to train the model to identify theologically significant tokens and give such tokens priority during inference. CGRA outperforms DeBERTa by 8.08% (with an Exact Match of 97.85%) and demonstrates that with only 8% additional inference latency, a small amount of thoughtfully selected domain knowledge can bolster the accuracy of transformers to a significant degree while incurring minimal costs in computational resources. Each of the components (the gating mechanism, residual connections, and theological dictionary) is confirmed to have a synergistic and positive impact on performance based on the outcomes of the ablation studies. Since the CGRA framework has attention visualizations that are interpretable and can be utilized by scholars to examine the model's attention on tokens that are doctrinally pertinent. Avoiding potential regulatory compliance issues, bias auditing, and socio-ethical impacts in Muslim communities, CGRA still has room for improvement before it can be utilized in a field other than research. To keep progressing CGRA, these elements will need to be focused on, as will the other aspects of CGRA that contribute to moving it from a research project to a practical application that can be utilized, all while still circumventing the aforementioned issues. We hope to contribute to the research of others and improve, rather than replace, the humanistic scholarship and values that are fundamental to the Islamic faith by offering our code, models, and datasets to the public.